\documentclass[11pt]{article}

\usepackage[final]{acl}

\usepackage{times}
\usepackage{latexsym}

\usepackage[T1]{fontenc}

\usepackage[utf8]{inputenc}

\usepackage{microtype}

\usepackage{inconsolata}

\usepackage{graphicx}

\usepackage{enumitem}

\usepackage{booktabs}
\usepackage{array}
\usepackage{multirow}
\usepackage{subcaption}

\newcommand{\projname}{MTRAG-UN}

\newcommand{\radbench}{$\textrm{RB}_\textrm{llm}$}
\newcommand{\ragasF}{RL$_\textrm{F}$}
\newcommand{\agg}{$\textrm{RB}_\textrm{alg}$}

\title{\projname: A Benchmark for Open Challenges\\in Multi-Turn RAG Conversations}

\author{Sara Rosenthal, Yannis Katsis, Vraj Shah, \\ \textbf{Lihong He, Lucian Popa, Marina Danilevsky} \\ 
 IBM Research, USA \\
  {\tt sjrosenthal@us.ibm.com} \\}

\begin{document}
\maketitle
\begin{abstract}

We present \projname, a benchmark for exploring open challenges in multi-turn retrieval augmented generation, a popular use of large language models. We release a benchmark of 666 tasks 
containing over 2,800 conversation turns across 6 domains with accompanying corpora. Our experiments show that retrieval and generation models continue to struggle on conversations with UNanswerable, UNderspecified, and NONstandalone questions and UNclear responses. Our benchmark is available at \url{https://github.com/IBM/mt-rag-benchmark}
\end{abstract}

\section{Introduction}


Seeking information continues to be a popular use case for Large Language Models (LLMs) \cite{wang2024understandinguserexperiencelarge}. Thus, Retrieval Augmented Generation (RAG), particularly in the multi-turn interactions of LLM chat interfaces \cite{li2025singleturnsurveymultiturninteractions},  remains an important research area. Several benchmarks have been released to evaluate model performance on such tasks  \cite{dziri-etal-2022-faithdial,10.1145/3626772.3657860,kuo2024radbenchevaluatinglargelanguage}. In particular, the recent MTRAG benchmark \cite{katsis2025mtragmultiturnconversationalbenchmark} focused on multi-turn information-seeking conversations, constituting 842 tasks in four domains. They reported several interesting findings that highlighted areas of improvement including unanswerable questions and later conversation turns. 

We pick up on these suggested areas by focusing on user goals that are not achievable via a single question-response\footnote{We use `question' to refer to any user utterance} exchange with an LLM.
We show this via a new benchmark, complementary to MTRAG, that focuses on: 



\begin{itemize}[noitemsep, topsep=0pt, leftmargin=*]
    \item UNanswerable Question - the user question is not answerable~\cite{katsis2025mtragmultiturnconversationalbenchmark}
    \item UNderspecified Question - the user question is ill-formed or ambiguous, lacking the information to determine a clear intent
    \item NONstandalone Question - the user question cannot be understood without the prior turns
    \item UNclear Response - 
    the user doesn't understand, or disagrees with the model answer and requires clarification 
\end{itemize}

\begin{figure}[t]
    \centering
    \includegraphics[width=\columnwidth]{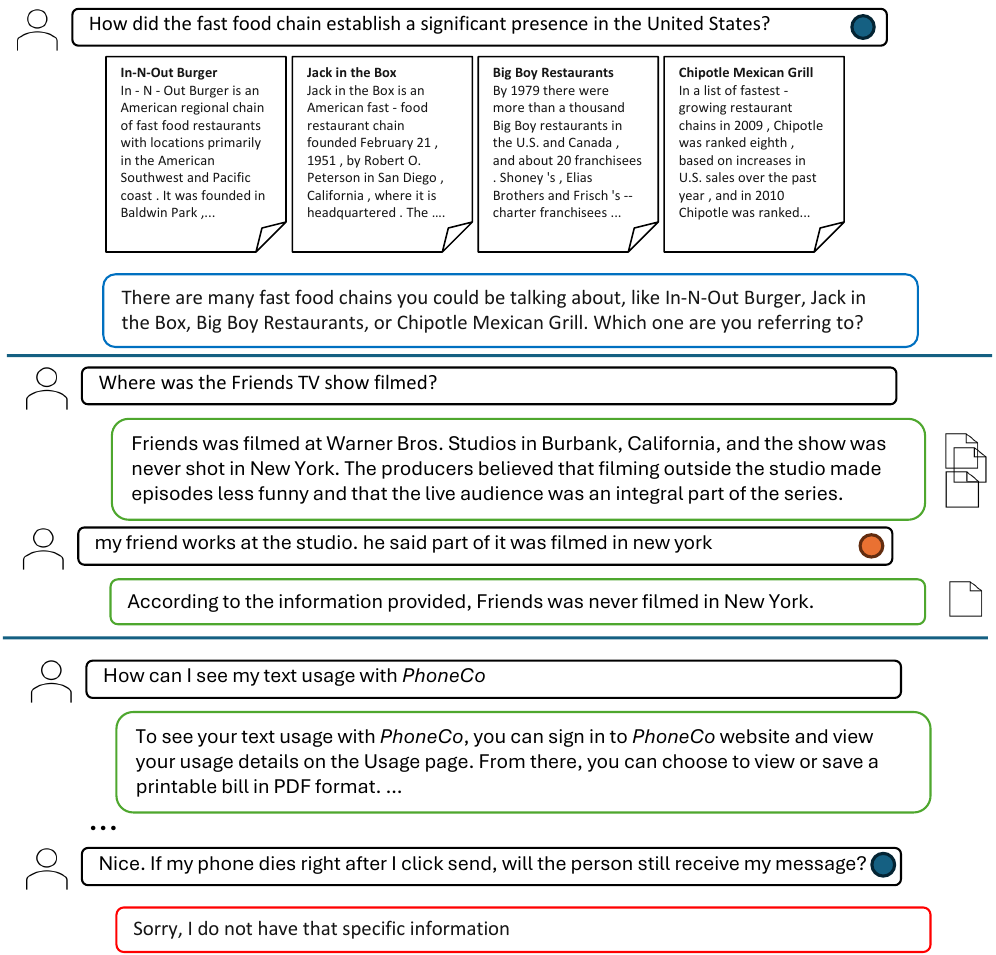}
    \caption{\small Portions of three conversations highlighting the challenges in \projname{}. The answerability is shown using the  assistant response color: 
    \textcolor{green}{answerable}, \textcolor{red}{unanswerable}, and \textcolor{blue}{underspecified}. 
    The multi-turn type is shown using the question circle: \textcolor{blue}{follow-up} and \textcolor{orange}{clarification}. The last two examples show non-standalone questions.}
    \label{fig:conversation_example}
\end{figure}

We thus refer to this new benchmark as \projname. An example of each task is shown in Figure~\ref{fig:conversation_example}. 
Our analysis shows that most frontier models struggle with handling such tasks, jumping to answer based on plausible but assumed interpretations of user intent. These challenges persist in both the retrieval and generation steps of multi-turn RAG.

\begin{figure*}[t]
    \centering
    \includegraphics[width=\textwidth]{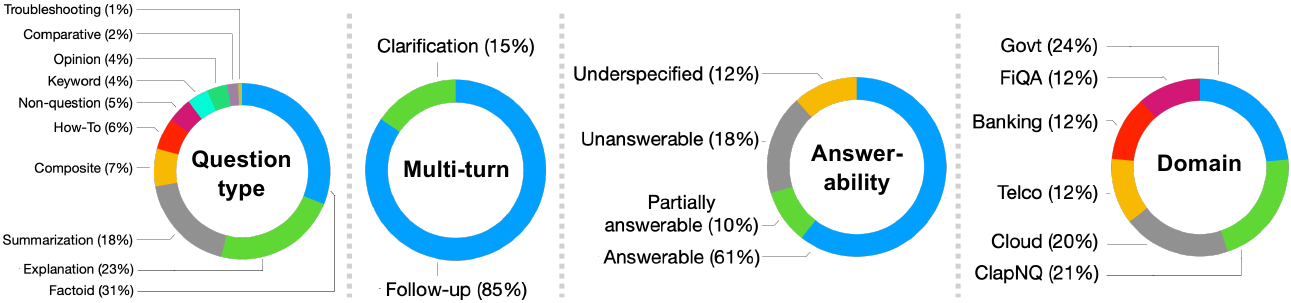}
\caption{\small Distribution of tasks in \projname\ based on different dimensions.}
\label{fig:combined-stats-by-dimension}
\end{figure*}

Our contributions are as follows:



\begin{itemize}[noitemsep, topsep=0pt,leftmargin=*]
    \item We present unexplored areas: UNanswerable, UNderspecified and NONstandalone user questions; and UNclear model responses.
    \item We add multi-turn conversations in two new corpora, Banking and Telco, to explore the use case of chatbots that are deployed in enterprise settings to support information-seeking questions
    \item We release \projname: A comprehensive benchmark consisting of 666 tasks for evaluating Retrieval, Generation, and the full RAG pipeline. The benchmark is available at: \url{https://github.com/IBM/mt-rag-benchmark} 
\end{itemize}

\section{Benchmark Creation}
\label{sec:benchmarkcreation}

We describe the tasks presented in \projname, as well as the document corpora used for the reference passages. The conversations were created by human annotators following the process described in \cite{katsis2025mtragmultiturnconversationalbenchmark}, using the \textsc{RAGaphene} platform~\cite{fadnis2025ragapheneragannotationplatform}. We collect a total of 666 human-generated conversations, with an average of 8 turns per conversation, and we describe the transformation of these conversations into the benchmark tasks at the end of this section.




\subsection{Task Definitions}
\label{sec:taskdefinitions}



\textbf{UNanswerable Question.} Such a question cannot be answered from retrieved passages, because no relevant passages could be found by the annotator. The MTRAG Benchmark~\cite{katsis2025mtragmultiturnconversationalbenchmark} showed that unanswerable questions are challenging for most LLMs. We ask annotators to include at least 2 unanswerable questions in each conversation, to ensure a sufficient and diverse data pool.


\textbf{UNderspecified Question.} A user question may be underspecified, ill-formed, or ambiguous, thus lacking enough information to determine a single clear intent. In such cases, rather than producing a wrong answer or replying with ``I don't know", the LLM agent should detect that the user question is unclear and get back to the user, either by pointing out missing details, presenting several plausible interpretations, or listing options based on the underlying passages. 
Conversations with underspecified questions were created 
via a combination of human and synthetic generation.
In the former, annotators were asked to write conversations that explicitly ended with an underspecified question. In the latter, an underspecified question (also written by a human) was stitched as a last turn on an existing human annotated multi-turn conversation. Relevant passages were added for the underspecified question using query expansion methods with a context relevance filter, in order to generate a rich set of passages to simulate the case of multiple interpretations. The reference response was generated using an LLM 
followed by human correction. 
The resulting conversations went through a careful human validation process. Appendix \ref{sec:clarification_appendix} gives further details. 


\textbf{NONstandalone Question.} In a multi-turn conversation, later turns can implicitly reference information in earlier turns. Such questions are considered non-standalone as they require the prior turns to be understood. We directed the annotators to include more non-standalone questions, an interesting challenge for retrieval.  

\textbf{UNclear Response (aka Clarification).} In a multi-turn conversation, a user may want to ask a clarification question if they don't clearly understand or disagree with the model answer to their previous question (e.g., “it was filmed in new york” in Figure \ref{fig:conversation_example}). Though the MTRAG benchmark included some clarification questions, these were not separately called out or evaluated. 

\begin{table}[t]
\centering
\small
\begin{tabular}{l r r r}
    \toprule
    {\bf Corpus} & {\bf Documents (D)} & {\bf Passages (P)} & {\bf Avg P/D}\\
    \midrule
    Banking & 4,497 & 33,380 & 7.4\\
    Telco & 4,616 & 52,350 & 11.3\\
    \bottomrule
\end{tabular}
\caption{\small Statistics of new document corpora in \projname.}
\label{tbl:corpora-stats}
\end{table}

\subsection{Document Corpora}


\projname\ consists of six document corpora: the original four corpora included in MTRAG (CLAPNQ \cite{rosenthal-etal-2025-clapnq}, FiQA \cite{macedo2018fiqa}, Govt, Cloud), and two new corpora from the domains of Banking and Telco (see Table \ref{tbl:corpora-stats}.) These new domains provide enterprise content, an unexplored area in 
MTRAG and other RAG benchmarks. Each of the corpora was created by crawling \textasciitilde{~1K} web-pages from several companies in the banking and telecommunications sector 
using seed-pages and crawling their neighborhood to ensure sets of inter-connected pages suitable for writing complex conversations on a given topic. 

\subsection{Benchmark: Tasks and Statistics}


From each conversation, we picked a single turn and created an evaluation task containing the entire conversation up to (and including) the question of the chosen turn, leading to the 666 evaluation tasks comprising the \projname\ benchmark. For conversations with underspecified questions, we chose the  turn containing the underspecified question. The remaining conversation turns were picked through a random process biased to give preference to challenging UN-turns. The resulting distribution of tasks is shown in Figure \ref{fig:combined-stats-by-dimension}. Compared to MTRAG, the \projname\ benchmark includes 6 instead of 4 domains, contains underspecified questions, has a higher representation of unanswerables/partially answerables (a combined 28\% vs 15\% in MTRAG), and a set of explicitly labeled clarification questions (15\% of the tasks). \projname\ is also biased against selecting the first turn of a conversation (8\% of the tasks - see Appendix, Figure \ref{fig:combined-stats-by-turn}) , which was found to be easier for LLMs \cite{katsis2025mtragmultiturnconversationalbenchmark}. 




\section{Evaluation}

We report retrieval and evaluation results on the \projname{} benchmark. Unless otherwise specified, all experiments and settings mimic the MTRAG paper ~\cite{katsis2025mtragmultiturnconversationalbenchmark}.

\begin{table}
\small
\centering
\begin{tabular}{c||l|l|l||l|l}
\toprule
&	& \multicolumn{2}{c||}{\bf Recall}			&	\multicolumn{2}{c}{\bf nDCG} \\
	&		&	@5	&	@10	&	@5	&	@10	\\
    \midrule
BM25	&	LT	&	0.29	&	0.38	&	0.27	&	0.31	\\
	&	RW	&	0.36	&	0.47	&	0.34	&	0.39	\\
    \midrule
BGE-base 1.5	&	LT	&	0.25	&	0.32	&	0.23	&	0.26	\\
	&	RW	&	0.38	&	0.49	&	0.35	&	0.40	\\
     \midrule
Granite R2	&	LT	&	0.29	&	0.38	&	0.28	&	0.32	\\
	&	RW	&	0.40	&	0.51	&	0.37	&	0.42	\\
     \midrule
Elser	&	LT	&	0.40	&	0.49	&	0.36	&	0.40	\\
	&	RW	&	0.49	&	0.60	&	0.45	&	0.51	\\
\bottomrule
\end{tabular}
\caption{\small Retrieval Performance using Recall and nDCG metrics for Last Turn (LT) and Query Rewrite (RW)}
\label{tab:retrieval-results}
\end{table}

\begin{table}[]
\small
    \centering
    \begin{tabular}{l||l|c|l}
        &  \textbf{Subset}	&	\textbf{LT}	&	\textbf{RW}	\\
         \midrule
\multirow{2}{*}{Standalone} & No (214)	&	0.39	&	0.52	\\
& Yes (254)	&	0.40	&	0.46	\\
    \end{tabular}
    \caption{Elser R@5 standalone results}
    \label{tab:retrieval-by}
\end{table}

\subsection{Metrics}



\begin{figure*}[t]
\centering
    \begin{subfigure}{.32\textwidth}
    \centering
    \includegraphics[width=.99\textwidth]{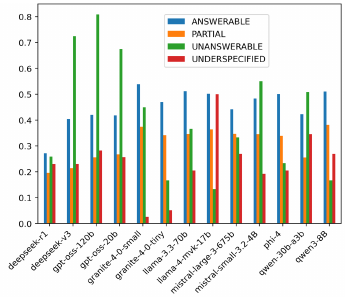}
    \caption{By question answerability}
    \label{fig:generation-breakdown-answerability}
    \end{subfigure}
    \begin{subfigure}{.33\textwidth}
    \centering
    \includegraphics[width=.99\textwidth]{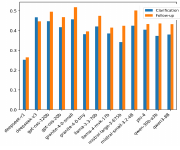}
    \caption{By multi-turn type}
    \label{fig:generation-breakdown-turn}
    \end{subfigure}
    \begin{subfigure}{.33\textwidth}
    \centering
    \includegraphics[width=.98\textwidth]{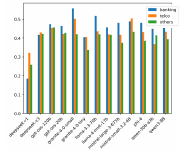}
    \caption{By domain}
    \label{fig:generation-breakdown-domain}
    \end{subfigure}
    \caption{\small Generation results in the Reference ($\bullet$) setting using, \agg, on three different dimensions.} 
    \label{fig:generation-breakdown}
\end{figure*}

We adopt the evaluation metrics of~\cite{katsis2025mtragmultiturnconversationalbenchmark}: (1) reference-based \radbench{} and \agg, (2) the IDK ("I Don't Know") judge, and (3) faithfulness judge from RAGAS \ragasF. All evaluation metrics are conditioned to account for answerability.
We use the open-source GPT-OSS-120B instead of the proprietary GPT-4o-mini as judge (Correlation is still aligned with human judgments - see Appendix~\ref{sec:appendix}). All other judges are consistent with those reported in MTRAG~\cite{katsis2025mtragmultiturnconversationalbenchmark}.
We create a new metric for the underspecified instances 
run with GPT-OSS-120b (See prompt in Appendix Figure~\ref{fig:clarify-judge}). Its accuracy on 80 random llama-4 and gpt-oss-120b model responses from underspecified instances is 96.2\%. These instances are not classified using the other metrics.

\subsection{Retrieval}


We ran retrieval experiments on the 468 answerable and partially answerable questions. We follow the experiments in MTRAG by running on lexical (BM25), sparse (Elser), and dense models. We added a newer SOTA dense embedding model, Granite English R2~\cite{awasthy2025graniteembeddingr2models}, and compare it to BGE-base 1.5~\cite{bge_embedding} as reported in the original paper. We also experimented with using newer open source models for Query Rewrite \cite{Sun2023} with the same prompt reported in the MTRAG paper and found that GPT-OSS 20B performed best. In all cases Query Rewrite outperforms the last turn. Granite English R2 performs better than BGE-base 1.5 embeddings, but Elser still performs best. The macro-average results across all domains are shown in Table~\ref{tab:retrieval-results}. We also provide a breakdown by standalone as provided in MTRAG in Table~\ref{tab:retrieval-by}. We have a considerably larger amount of non-standalone questions that require rewrite (17.7\% in MTRAG and 45.7\% in \projname{}). Rewrite helps for both standalone and non-standalone questions, but more so for non-standalone questions. 

Our new domains of Banking and Telco perform worse than the other domains with .32 and .39 R@5 respectively (compared to an average of .52 R@5 for the other domains). 
To investigate this gap, we analyzed corpus-level characteristics and found that Banking and Telco contain substantially longer documents and denser hyperlink structures, suggesting stronger cross-page dependencies typical of enterprise web content. 
Additionally, these domains include multiple companies with structurally similar pages (e.g., checking accounts or credit card offers), which likely increases retrieval difficulty due to content similarity across sources.
Overall, our scores are lower than MTRAG, highlighting that more work is needed for multi-turn retrieval.

\begin{table}[!t]
\centering
\small
\setlength{\tabcolsep}{3pt}
\begin{tabular}{lcc|cc|cc}
\toprule
&  \multicolumn{2}{c}{\bf \ragasF}  &  \multicolumn{2}{c}{\bf \radbench}  &  \multicolumn{2}{c}{\bf \agg} \\
& $\bullet$ & $\circ$ & $\bullet$ & $\circ$ & $\bullet$ & $\circ$ \\
\midrule
target & 0.85 & 0.69 & 0.96 & 0.92 & 0.89 & 0.88 \\
gpt-oss-120b & \textbf{0.65} & \underline{0.59} & \textbf{0.76} & \textbf{0.65} & \textbf{0.46} & \underline{0.37} \\
gpt-oss-20b & 0.60 & 0.55 & 0.67 & \underline{0.63} & 0.43 & 0.36 \\
deepseek-v3 & \underline{0.63} & \textbf{0.60} & 0.61 & 0.58 & 0.42 & \underline{0.37} \\
deepseek-r1 & 0.47 & 0.46 & 0.54 & 0.52 & 0.26 & 0.23 \\
granite-4-0-small & 0.62 & 0.56 & 0.55 & 0.53 & \underline{0.45} & \textbf{0.38} \\
granite-4-0-tiny & 0.48 & 0.46 & 0.50 & 0.50 & 0.35 & 0.31 \\
qwen-30b-a3b & 0.61 & \textbf{0.60} & \underline{0.68} & 0.60 & 0.41 & 0.36 \\
qwen3-8B & 0.57 & 0.55 & 0.64 & 0.58 & 0.41 & 0.36 \\
llama-4-mvk-17b & 0.62 & 0.58 & 0.59 & 0.57 & 0.42 & \underline{0.37} \\
llama-3.3-70b & 0.62 & 0.58 & 0.58 & 0.55 & 0.43 & \textbf{0.38} \\
mistral-small 24b & \underline{0.63} & \underline{0.59} & 0.67 & 0.57 & \underline{0.45} & \underline{0.37} \\
mistral-large 675b & 0.55 & 0.52 & 0.67 & 0.60 & 0.39 & 0.34 \\
phi-4 & 0.54 & 0.49 & 0.64 & 0.57 & 0.40 & 0.34 \\
\bottomrule
\end{tabular}
\caption{\small Generation by retrieval setting: Reference ($\bullet$) and RAG ($\circ$). The best result is \textbf{bold} and runner-up is \underline{underlined}.}
\label{tbl:judge_results_by_RAGsetting}
\end{table}

\subsection{Generation}

We ran generation experiments using the original prompt used in MTRAG \cite{katsis2025mtragmultiturnconversationalbenchmark} with an additional sentence to
to accommodate the possibility of underspecified questions:

\noindent\fbox{
    \parbox{.98\columnwidth}{
\scriptsize
Given one or more documents and a user question, generate a response to the question using less than 150 words that is grounded in the provided documents. If no answer can be found in the documents, say, "I do not have specific information". If a question is underspecified — e.g., it has multiple possible answers, a broad scope, or needs explanation — include that further clarification/information is needed from the user in your response.
}}
\\

In the reference task 
we send up to the first 10 relevant passages for generation. In the RAG task, we send the top 5 retrieved passages using Elser with query rewrite. 

Table~\ref{tbl:judge_results_by_RAGsetting} presents the generation evaluation results for both reference and RAG settings. We evaluate a diverse set of LLMs, including GPT-OSS~\cite{openai2025gptoss}, DeepSeek-V3~\cite{deepseekv3}, DeepSeek-R1~\cite{deepseekr1}, Granite-4~\cite{granite4_2025}, Qwen3~\cite{qwen3_2025}, Llama~\cite{llama4_2025, llama3_2024}, Mistral~\cite{mistralai_2025}, and Phi-4~\cite{phi_2024}. Model scores remain significantly lower than target answer scores, indicating room for improvement in multi-turn RAG. Larger models usually perform better within each model family, and performance in the reference setting is consistently higher than RAG, reflecting the added difficulty introduced by retrieval noise. GPT-OSS-120B achieves the best scores, while DeepSeek-V3, Qwen-30B and Mistral-Small-24B remain competitive.

Figure~\ref{fig:generation-breakdown} shows the generation quality by different dimensions: answerability, multi-turn type, and domain. 
While most models perform worse on unanswerables, DeepSeek-V3 and GPT-OSS models exhibit comparatively robust behavior by frequently responding with IDK. This is a stark improvement over the takeaways from prior work~\cite{katsis2025mtragmultiturnconversationalbenchmark}, where no models handled unanswerables well. Performance on underspecified question is consistently low, as models are generally eager to answer based on a plausible but assumed interpretation of the question.
Clarification questions show lower performance than follow-up questions. This suggests that current models are better at conversational continuation than for intent refinement and self-correction. We find that the performance across the two new domains is largely comparable, while the other domains (average performance reported in Figure~\ref{fig:generation-breakdown-domain}) trend lower due to the challenging FiQA corpus~\cite{katsis2025mtragmultiturnconversationalbenchmark}.

\section{Conclusion and Future Work}

The \projname{} benchmark of 666 tasks and baseline results provided in our paper highlight existing and ongoing challenges in multi-turn RAG. We release our benchmark\footnote{\url{https://github.com/IBM/mt-rag-benchmark}} to encourage advances in this important topic. In the future, we plan to release multilingual RAG conversations.

\section{Acknowledgments}

We would like to thank our annotators for their high-quality work in generating and evaluating this dataset: Mohamed Nasr, Joekie Gurski, Tamara Henderson, Hee Dong Lee, Roxana Passaro, Chie Ugumori, Marina Variano, and Eva-Maria Wolfe.

\section*{Limitations}

Our conversations are limited to English and 6 closed domains. They are created by a small set of human annotators and thus likely contain biases toward those individuals and Elser retriever and the Mixtral 8x7b generator used to retrieve passages and generate the initial response respectively. Expanding the annotator pool and creating conversations in other languages would improve these limitations. 

\bibliography{custom}

\appendix

\section{Stats and Metrics}
\label{sec:appendix}

\begin{figure}[t]
    \centering
    \includegraphics[width=0.6\columnwidth]{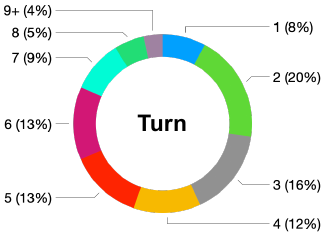}
\caption{\small Distribution of tasks in \projname\ based on conversational turn.}
\label{fig:combined-stats-by-turn}
\end{figure}

A distribution of tasks by turn is provided in Figure~\ref{fig:combined-stats-by-turn}. \projname\ does not include conversational questions (e.g., ``Hi", ``Thank you"), since, as noted in MTRAG (which included them in the benchmark but not in the evaluation), more work is required to develop appropriate evaluation metrics for them.

\begin{figure}[t]
    \centering
    \begin{subfigure}{.25\textwidth}
    \centering
    \includegraphics[width=.98\textwidth]{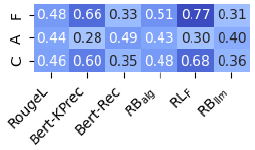}
    \caption{With \emph{Faithfulness (F)}, \emph{Appropriateness (A)}, and \emph{Completeness (C)}.}
    \label{fig:mtrag1-correlation-fig1}
    \end{subfigure}
    %
    \begin{subfigure}{.22\textwidth}
    \centering
    \includegraphics[width=.98\textwidth]{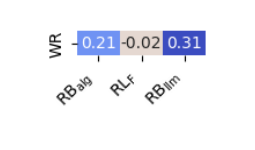}
    \caption{With Win-Rate (WR)}
    \label{fig:mtrag1-correlation-fig2}
    \end{subfigure}
    \caption{Weighted Spearman correlation: automated judge metrics vs human evaluation metrics.}
    \label{fig:mtrag1-correlation}
\end{figure}

To ensure that using GPT-OSS-120B in place of the GPT-4o-mini as the judge does not negatively affect the quality of the evaluation results, we repeated the correlation analysis of ~\cite{katsis2025mtragmultiturnconversationalbenchmark} using the open-source model as the judge. The results are depicted in Figure~\ref{fig:mtrag1-correlation}. We observe that the correlation between the judges and the humans judgments using the open source model improved slightly or remained consistent compared to using the proprietary model as the judge.

\section{Details on UNderspecified}
\label{sec:clarification_appendix}

\begin{figure}[t]
    \centering
    \includegraphics[width=1.0\columnwidth]{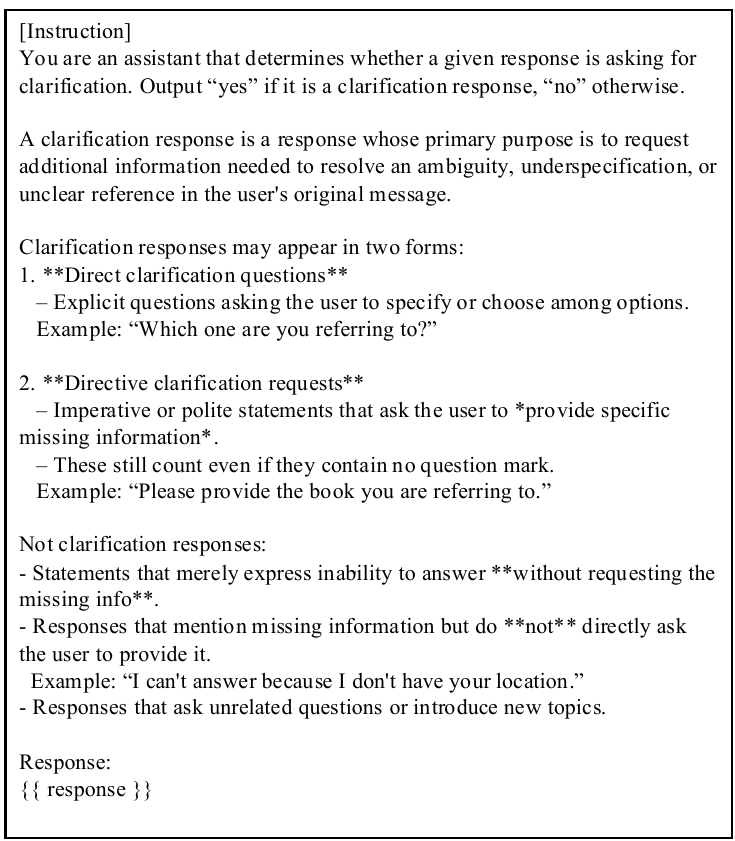}
\caption{Prompt used for clarification judge.}
\label{fig:clarify-judge}
\end{figure}

Figure~\ref{fig:conversation_example} shows an example of a conversation where the last user turn is an underspecified question (asking about a vague fast food chain in the US), together with a set of reference passages from the corpus, and a target response for what the model should ask back from the user. The patterns for the model response follow three general categories, each ending with a request to the user to give more information (see also Table~\ref{tbl:clarification-pattern}):  

\begin{enumerate}
\item  Hedging with answers (for the case with few options -- e.g., 2-3):  list the few options and provide a brief description or answer associated with each.

 \item Hedging over list (for the cases with medium number of options -- e.g., 4-8): an enumeration of the plausible options without additional explanatory content.

\item Open-domain (for the cases where there are many/unbounded options): directly ask the user for disambiguation over the {\em type} of entity that they may have in mind. 
\end{enumerate}


\begin{table}[h]
\centering
\scriptsize
\begin{tabular}{l p{0.6\columnwidth}}
\toprule
\textbf{Pattern} & \textbf{Example} \\
\midrule
Hedging with answer & There are many astronauts you could be referring to, such as Ellen Ochoa, who was the first Hispanic woman to go to space and has received numerous awards, including the Presidential Medal of Freedom, or Kalpana Chawla, who was the first woman of Indian origin to go to space and tragically died in the Columbia disaster in 2003. Which one are you talking about?
  \\
\midrule
Hedging over list & There are many fast food chains you could be talking about, like In-N-Out Burger, Jack in the Box, Big Boy Restaurants, or Chipotle Mexican Grill. Which one are you referring to?  \\
\midrule
Open-domain  & Which modern smart design segment are you talking about?  \\
\bottomrule
\end{tabular}
\caption{Types of response to underspecified questions.}
\label{tbl:clarification-pattern}
\end{table}


\subsection{Stitching of the underspecified questions}

In Section~\ref{sec:taskdefinitions}, we described underspecified questions written by a human that were stitched as a last turn onto an existing human annotated multi-turn conversation. 
The process we implemented was a very controlled one, where stitching was done in two ways: a) by finding existing conversations on the same or very similar topic, simulating the case where the new turn is not out of place (75\% of the underspecified tasks), and b) by finding existing conversations on a different topic, so that the new turn reflects a topic switch by the user, while still being an underspecified question (25\% of the underspecified tasks). The second case could change the flow of the conversation, but we believe that it adds an additional challenge to the models evaluated on such data. 
In particular, it reflects the realistic scenario where users change topics sometimes randomly, but we still want the models to be able to detect that and react accordingly. 

\subsection{Validation}

The underspecified questions went through careful validation including filtering (e.g., of cases where the last turn intent would accidentally become clear because the addition of the context in which it is being stitched onto), editing of the last turn or of the reference model response to it, and most of the time 
just plain validation.

\end{document}